\documentclass{article}

\usepackage[preprint]{neurips_2023}

\usepackage[utf8]{inputenc} 
\usepackage[T1]{fontenc}    

\usepackage{graphicx}
\usepackage{amsmath}
\usepackage{amssymb}
\usepackage{tabularx}
\usepackage{cite}
\usepackage{amsfonts}
\usepackage{algorithmic}
\usepackage{subcaption}
\usepackage{multirow}
\usepackage{booktabs}       

\usepackage{bm}
\usepackage{color}
\usepackage{xcolor}
\usepackage{algorithm}
\usepackage{xspace}
\usepackage{soul}

\usepackage{enumitem}
\usepackage{textcomp}  
\usepackage{scalerel}  
\usepackage{wrapfig}

\usepackage[colorlinks,linkcolor=blue]{hyperref}
\usepackage{cleveref}
\crefname{section}{Sec.}{Secs.}
\Crefname{section}{Section}{Sections}
\Crefname{table}{Table}{Tables}
\crefname{table}{Tab.}{Tabs.}
\newcommand*\samethanks[1][\value{footnote}]{\footnotemark[#1]}
\def\logo{\scalerel*{\includegraphics{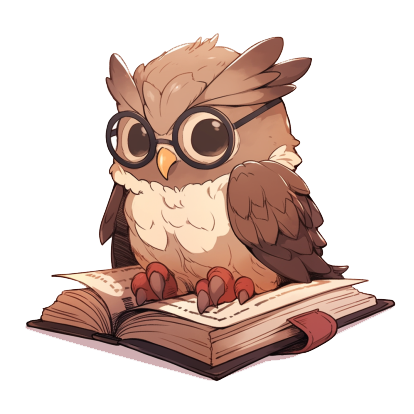}}{\textrm{\textbigcircle}}}
\newcommand{\modelname}{mPLUG-DocOwl\xspace}
\newcommand{\evalname}{LLMDoc\xspace}
\newcommand{\titledmodelname}{\modelname\logo\xspace}

\title{\titledmodelname: Modularized Multimodal Large Language Model for Document Understanding}

\author{
Jiabo Ye\thanks{Equal contribution}\hspace{1.5mm}, Anwen Hu\samethanks[1], Haiyang Xu\thanks{Corresponding author}, Qinghao Ye, Ming Yan\samethanks[2], Yuhao Dan,  \\ \textbf{Chenlin Zhao, Guohai Xu, Chenliang Li, Junfeng Tian, Qian Qi, Ji Zhang, Fei Huang}\\
DAMO Academy, Alibaba Group \\
{\small \texttt{\{yejiabo.yjb, huanwen.haw, shuofeng.xhy, yeqinghao.yqh, ym119608\}@alibaba-inc.com}}
}

\begin{document}

\maketitle

\begin{abstract}

Document understanding refers to automatically extract, analyze and comprehend information from various types of digital documents, such as a web page. Existing Multi-model Large Language Models (MLLMs), including mPLUG-Owl, have demonstrated promising zero-shot capabilities in shallow OCR-free text recognition, indicating their potential for OCR-free document understanding. Nevertheless, without in-domain training, these models tend to ignore fine-grained OCR features, such as sophisticated tables or large blocks of text, which are essential for OCR-free document understanding. In this paper, we propose \modelname based on mPLUG-Owl for OCR-free document understanding. Specifically, we first construct a instruction tuning dataset  featuring a wide range of visual-text understanding tasks. Then, we strengthen the OCR-free document understanding ability by jointly train the model on language-only, general vision-and-language, and document instruction tuning dataset with our unified instruction tuning strategy. We also build an OCR-free document instruction understanding evaluation set \evalname to better compare models' capabilities on instruct compliance and document understanding. Experimental results show that our model outperforms existing multi-modal models, demonstrating its strong ability of document understanding. Besides, without specific fine-tuning, \modelname generalizes well on various downstream tasks. Our code, models, training data and evaluation set are available at https://github.com/X-PLUG/mPLUG-DocOwl. 

\end{abstract}

\section{Introduction}
Large language models (LLMs) like ChatGPT~\citep{chatgpt}, BLOOM~\citep{bloom}, and LLaMA~\citep{llama} have undergone rapid development to enable the realization of general artificial intelligence, boasting impressive zero-shot capabilities across diverse linguistic applications. With the LLM as the language decoder, Multimodal large language models (MLLMs) such as MiniGPT-4~\citep{minigpt4}, LLaVA~\citep{llava}, and mPLUG-Owl~\citep{mplugowl} have demonstrated remarkable zero-shot performance in various open-ended vision-and-language tasks. 
These models are trained to align text and images during the pre-training phase, and then to promote diverse abilities during the instruction tuning phase. 
Interestingly, these MLLMs exhibit superficial OCR-free text recognition abilities without explicit training on visual text understanding datasets~\citep{mplugowl,llmocr}. Nevertheless, due to lacking specific training, these models still face the challenge of comprehending intricate relationships between visual text and objects in diverse types of images, such as charts, documents and webpages.

By performing unified instruction tuning for Document Understanding upon the mPLUG-Owl~\citep{mplugowl}, we further propose a modularized MLLM~\citep{mplug, mplug2}, namely \modelname.
Our approach utilizes a modularized framework similar to mPLUG-Owl \citep{mplugowl}, which incorporates a visual abstractor module to link a pre-trained LLM with a visual knowledge module, achieving the alignment of text and images. To enhance diverse document understanding capabilities, we reorganize various downstream document understanding tasks in the same form of instructions. To maintain general uni/multi-modal abilities, we also include language-only and general vision-and-language instruction datasets used by mPLUG-Owl to train the \modelname. During training, both the visual knowledge module and LLM decoder are frozen, only the visual abstractor and the Low-Rank Adaption (LoRA)~\citep{lora} in LLM are fine-tuned. 

\modelname achieves ocr-free state-of-the-art performance on multiple commonly used document understanding datasets. Furthermore, our experiments on a carefully-built document instruction understanding evaluation set \evalname shows that \modelname achieves significantly better visual text understanding performance on various domains than existing MLMMs.

Our main contributions can be highlighted as follows:
\begin{itemize}
    \item We propose a modularized MLLM, \textbf{\modelname}, which is the first one to balance language-only, general vision-and-language, and document understanding based on unified instruction tuning.
    \item We carefully construct an instruction understanding test set with human evaluation, dubbed \textbf{\evalname}, to assess diverse  document understanding capabilities.
    \item Empirical results demonstrate that our \modelname surpasses existing methods on ocr-free document understanding, including multiple standard benchmarks and \evalname.
\end{itemize}

\section{Related Work}
\subsection{Visual Text Understanding}
There are two types of models for understanding images that contain rich textual information. The first kind of approaches~\citep{layoutlm,layoutlmv3,qctextcap,udop,tap} utilize off-the-shelf OCR models or APIs to recognize text from images, and then design pretraining tasks to facilitate cross-modality alignment between visual and textual inputs. On the other hand, end-to-end approaches~\citep{dessurt, donut,pix2struct} utilize a high-resolution image encoder to learn text recognition during the pretraining stage. Both two types of models rely on specific finetuning on different downstream datasets and can't achieve open-domain instruction understanding performance like Multimodal Large Language Models.

\subsection{Multimodal Large Language Model}
Large Language Models (LLMs) have demonstrated impressive zero-shot abilities across various open-ended tasks. Recent research has also explored the application of LLMs for multi-modal generation, utilizing two different paradigms: systematic collaboration and end-to-end trained models. Systematic collaboration approaches, such as Visual ChatGPT \citep{visualchatgpt} and MM-REACT \citep{mmreact}, leverage various vision experts or tools to express visual information with text descriptions. Subsequently, LLMs, such as ChatGPT \citep{chatgpt}, can act as agents and select appropriate experts and tools for visual understanding. Finally, LLMs would summarize the output of these experts to answer user queries. On the other hand, some approaches, such as MiniGPT-4 \citep{minigpt4}, LLaVA \citep{llava}, and mPLUG-Owl \citep{mplugowl}, leverage LLMs to build unified models for multi-modality with limited connected parameters. These methods show superficial OCR-free text recognition abilities under the zero-shot setting. However, for complicated document understanding,  due to lacking in-domain training,  they encounter challenges in handling diverse image types, recognizing rich texts and comprehending  relationships between visual semantic and text information. In this work, through unified instruction tuning, \modelname achieves much better document understanding performance and maintains general uni/multi-modal abilities.

\section{\modelname}
\begin{figure}[tp]
    \centering
    \includegraphics[width=1.0\linewidth]{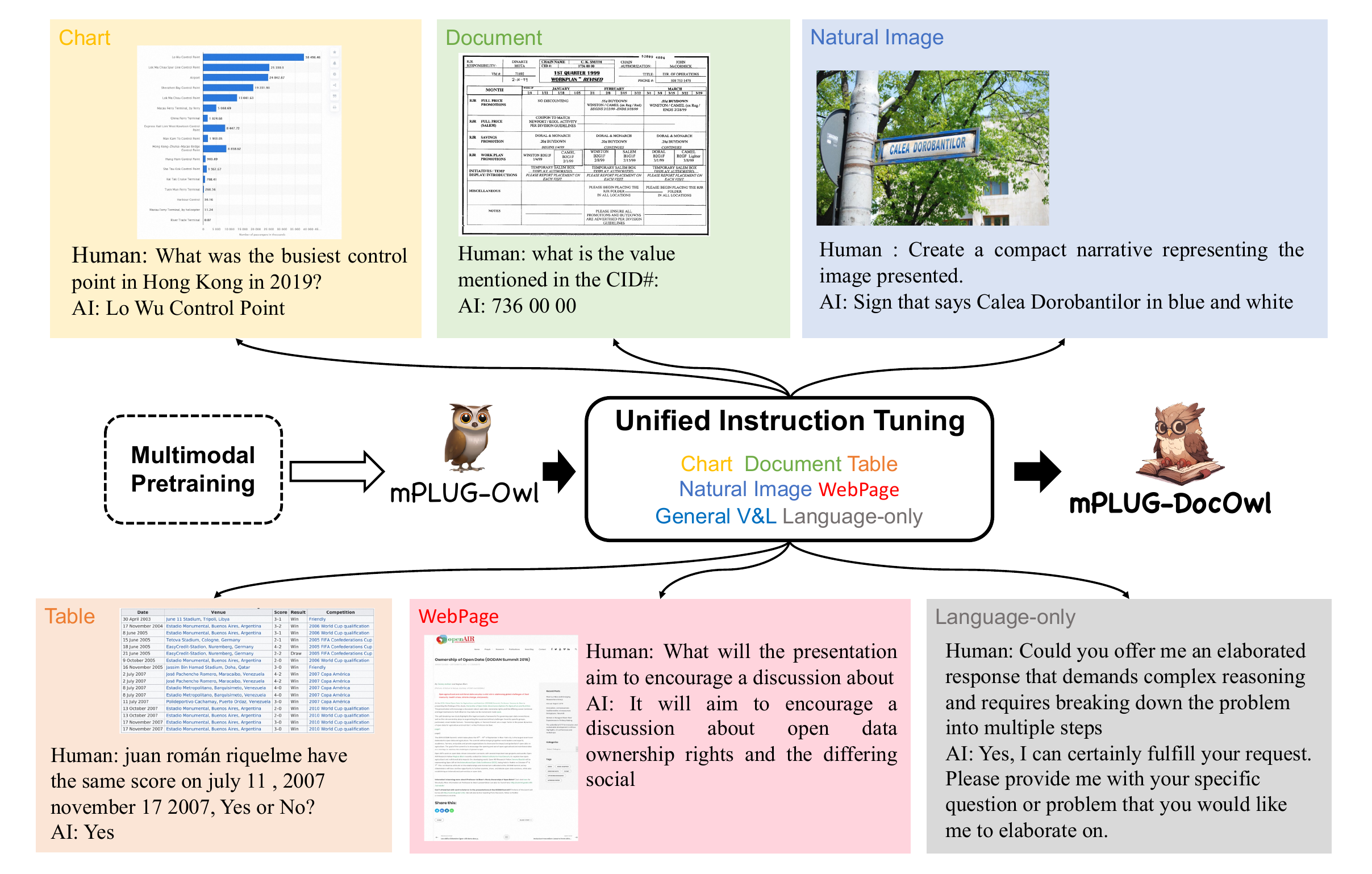}
    \caption{The summary of the instruction tuning paradigm of our \modelname.}
    \label{fig:overall_model}
\end{figure}

\subsection{Architecture}
The architecture of \modelname is based on a popular multi-modal language model, mPLUG-Owl \citep{mplugowl}, which comprises a pre-trained visual foundation model, a visual abstractor, and a language foundation model. The visual foundation model is responsible for extracting visual features from the input images, and the visual abstractor distills these features using a set of learnable tokens. The resulting visual features are then concatenated with the word embeddings of the input sentence and fed into the language model to generate the response. This powerful architecture allows for accurate and efficient multi-modal language processing.

The mPLUG-Owl \citep{mplugowl} exhibits superficial OCR ability when presented with images containing salient text. Inspired by this, we propose to further fine-tune the model with document instruction tuning data for better document understanding performance, covering document, table, chart and natural image and webpage. During fine-tuning, we freeze the visual encoder and the language model and train the visual abstractor. We also adopt the low-rank adaptation approach (LoRA)~\citep{lora} to enhance the language model's ability.

\subsection{Instruction Tuning Data}

This section introduces the composition of our instruction tuning data in detail. To ensure the versatility of \modelname, we collect diverse document understanding datasets with different task formats, including Visual Question Answering (VQA) \citep{vqa}, Information Extraction (IE), Natural Language Inference (NLI) \citep{NLI}, and Image Captioning (IC). mPLUG-Owl~\citep{mplugowl} performs instruction tuning with a unified format as "<image>Human:\{question\} AI:\{answer\}". In this work, we convert different document understanding tasks to the same format as mPLUG-Owl \citep{mplugowl} by replacing the \{question\} and \{answer\} placeholders as follows.

\noindent{\textbf{Visual Question Answering}} We simply use the raw question and answer as the \{question\} and \{answer\} placeholders. We collect VQA datasets on diverse domains, including ChartQA~\citep{chartqa}, DocVQA~\citep{docvqa}, InfographicsVQA (InfoVQA)~\citep{infovqa}, WikiTableQuestions (WTQ)~\citep{wikitableqa}, TextVQA \citep{textvqa} and VisualMRC~\citep{visualmrc}.

\noindent{\textbf{Information Extraction}} requires the model to extract key-value pairs from the input image. The `keys' (or `categories') are always a stationary set. To convert this task to the instruction tuning format, we treat the value as the \{answer\} and construct the \{question\} as `What is the value for the \{key\}?'. When the key does not exist in the image, the \{answer\} is set to `None'. We collect Information Extraction data from DeepForm~\citep{deepform}, and Kleister Charity (KLC)~\citep{klc}. 

\noindent{\textbf{Natural Language Inference}} is a binary classification task with labels `Entailed' and `Refuted'. Given a statement, we construct the \{question\} as `\{statement\}, Yes or No?'. The \{answer\} is `Yes' or `No' and refers to `Entailed' or `Refuted', respectively. TabFact~\citep{TabFact}, a natural language inference dataset about tables, is chosen for instruction tuning.

\noindent{\textbf{{Image Captioning}} aims to briefly describe an image with fluent language. We treat the caption as the \{answer\} and randomly choose a prompt as the \{question\} like LLaVa \citep{llava}. TextCaps~\citep{textcaps} is an appropriate captioning dataset on natural images with texts.

\paragraph{Language-only and General Vision-and-language Instruction Tuning.} To enhance the model's ability of language comprehension and multi-modal open-ended conversation, we follow mPLUG-Owl \citep{mplugowl} to introduce language-only and general vision-and-language instruction tuning data \citep{alpaca, vicuna,baize, llava}. 

\begin{figure}[tp]
    \centering
    \includegraphics[width=1.0\linewidth]{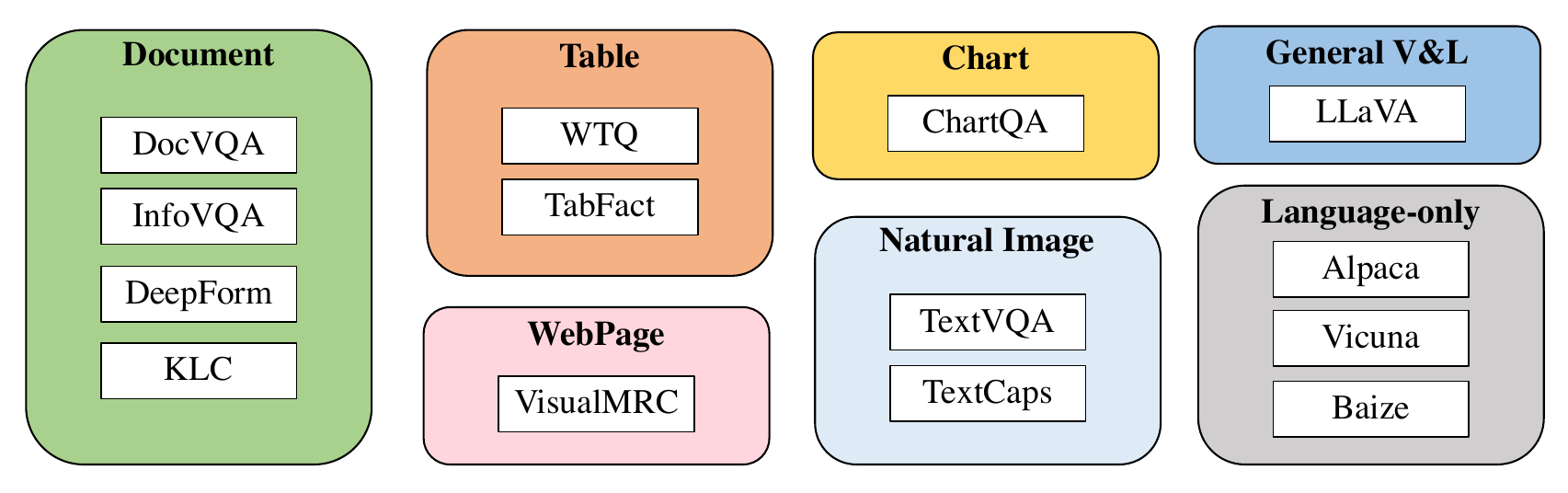}
    \caption{Different types of datasets used to train \modelname.}
    \label{fig:data}
\end{figure}

\Cref{fig:data} shows the composition of our instruction tuning data grouped by the dataset type. We use training sets of these datasets as instruction tuning data and evaluate models on test sets. 

\subsection{Training Details}
We adopt a two-stage training paradigm, where the Vision Transformer and Language model are kept frozen. In the first stage, both the visual abstractor and LoRA \citep{lora} in the language model are fine-tuned. The first stage only uses the document understanding data and takes 10 epochs. In the second stage, we further freeze the visual abstractor and only train the LoRA. Besides document understanding data, the language-only and general vision-and-language instruction tuning data are further introduced at this stage and up-sampled 6 times. The second stage takes 3 epochs. Other training hyper-parameters are the same as mPLUG-Owl \citep{mplugowl}.
\section{Experiment}

\subsection{LLMDoc}
Existing benchmarks are hard to evaluate the open-ended instruction understanding results given by MLMMs. For better compare the instruction understanding performance in the document domain, we further construct a test set with human evaluation, namely \evalname.

\paragraph{Data Collection}
To comprehensively evaluate the model's abilities, we consider five scenarios to construct our evaluation dataset, including table (TabFact \citep{TabFact}), chart (ChartQA \citep{chartqa}), document (DocVQA \citep{docvqa}), natural image (TextVQA \citep{textvqa}) and webpage (VisualMRC \citep{visualmrc}). Specifically, for each dataset, we sample 20 images from the test split. For 10 of these images, we adopt a raw question as the instruction. While for the other 10, we ask annotators to write instructions requiring stronger 
 capabilities like summarization, inference, and calculation. In total, we obtain 100 test samples.
\paragraph{Human Evaluation}
Following the rating criteria proposed in Self-Instruct~\citep{self-instruct}, we perform the human evaluation to score the model’s responses, where A > B > C > D and A represents `correct and satisfying response', B means `acceptable response with minor imperfections', C refers to `response to the instruction but has significant errors' and D means `irrelevant or invalid response'.

\begin{wrapfigure}{r}{0.5\linewidth}
    \centering
    \setlength{\abovedisplayskip}{1pt} 
    \setlength{\belowdisplayskip}{1pt} 
    \vspace{-12pt}
    \includegraphics[width=\linewidth]{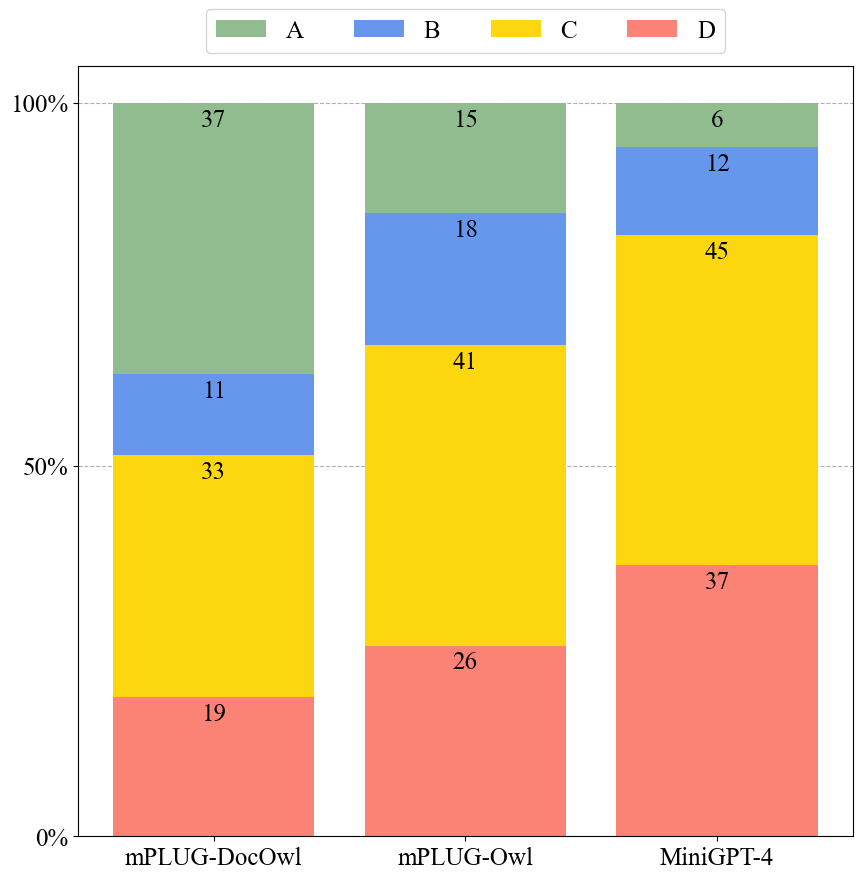}
    \caption{Human evaluation of \modelname, mPLUG-Owl and MiniGPT-4 on \evalname.}
    \label{fig:llm_comp}
    \vspace{-12pt}
\end{wrapfigure}

We compare \modelname with other popular mult-modal large language models, including mPLUG-Owl~\citep{mplugowl} and Mini-GPT4~\citep{minigpt4}, on \evalname. As shown in \Cref{fig:llm_comp}, \modelname achieves significantly better performance, with 37 responses being scored as ``A'', demonstrating the stronger understanding ability of \modelname in diverse document scenarios. Besides, it's worth noting that all models have some responses scored as ``C'' or ``D'', showing that instruction understanding performance in the document domain is still far from promising and needs more endeavor.

\subsection{Benchmark Evaluation}
Besides human evaluation, we also compare our \modelname with ocr-free 
 state-of-the-art document understanding models on public datasets. \Cref{tab:due_eval} shows the comparison with Dessurt~\citep{dessurt}, Donut~\citep{donut} and Pix2Struct~\citep{pix2struct} on DUE-Benchmark~\citep{due}, which mainly requires the text recognition and layout understanding abilities on documents and tables. Besides, \Cref{tab:other_eval} presents the evaluation on the chart, natural image and webpage datasets, which ask stronger ability to relate visual semantics and text information. Without finetuning on each dataset, our \modelname achieves comparable or even better performance.


\begin{table*}
    \caption{Comparison with ocr-free methods on DUE-Benchmark.}
    \label{tab:due_eval}
    \footnotesize
    \centering
    \begin{tabular}{c|cccccc}

    \toprule
    \textbf{Model}  & \textbf{DocVQA} & \textbf{InfoVQA} & \textbf{DeepForm} & \textbf{KLC} & \textbf{WTQ} & \textbf{TabFact} \\
    \midrule
    Dessurt &  63.2 & -& - & - & - & -  \\ 
    Donut  &  67.5 & 11.6 & \textbf{61.6} & 30.0 & 18.8 & 54.6 \\
    Pix2Struct$_{base}$ & 72.1 & 38.2 &- & - & - & - \\ 
    \midrule
    \modelname &  62.2 & \textbf{38.2} & 42.6 & \textbf{30.3} & \textbf{26.9} & \textbf{60.2} \\
    \bottomrule
    \end{tabular}
\end{table*}

\begin{table*}
    \caption{Comparison with ocr-free methods on chart, natural image and webpage understanding.}
    \label{tab:other_eval}
    \footnotesize
    \centering
    \begin{tabular}{c|cccc}
    \toprule
    \textbf{Model} & \textbf{ChartQA} & \textbf{TextVQA} & \textbf{TextCaps} & \textbf{VisualMRC} \\
    \midrule
    Donut & 41.8 & 43.5 & 74.4 & 93.91 \\
    Pix2Struct$_{base}$ & 56.0 & -& 88.0 & -  \\ 
    \midrule
    \modelname & \textbf{57.4} & \textbf{52.6} &\textbf{111.9} &\textbf{188.8} \\
    \bottomrule
    \end{tabular}
\end{table*}

\subsection{Qualitative Analysis}
\begin{figure}[tp]
    \centering
    \includegraphics[width=1.0\linewidth]{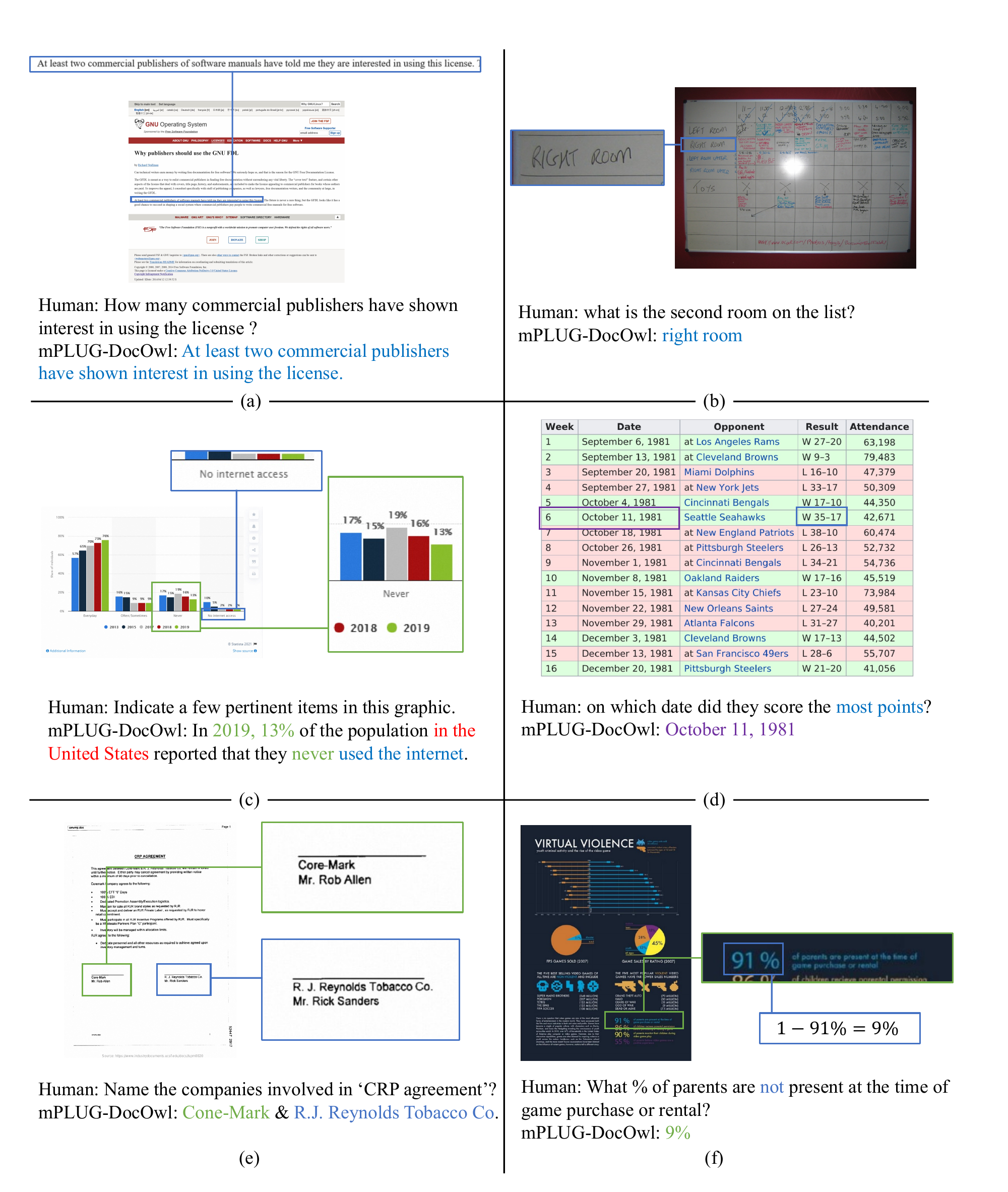}
    \caption{Qualitative results of \modelname.  The crucial regions and corresponding words are annotated with the same colors for clearer visualization. Wrong answers are colored \textcolor{red}{red}.}
    \label{fig:cases}
\end{figure}

\paragraph{Benchmark Results.} Qualitative results on different types of images are shown in \Cref{fig:cases}. Crucial regions and corresponding responses are annotated with the same colors. 
Case (a) shows that \modelname can accurately find the answer from a webpage screenshot with complex contents. Case (b) shows that \modelname is even able to understand hand-drawn tables and correctly recognize handwritten fonts. In case (c), \modelname can summarize key points from a chart. It successfully understands that the table is about internet usage and infers that ``Never'' means ``Never used internet''. However, it also generates illusory outputs, such as "in the United States". The question in case (d) requires the model to understand the ``Result'' column, compare the points and return the date with the best results. Case (e) demonstrates that our model is capable of processing scanned documents and distinguishing company and person names. Case (f) shows that \modelname can not only recognize small and blurry text but also perform simple calculations following the user intent.

\begin{figure}[tp]
    \centering
    \includegraphics[width=1.0\linewidth]{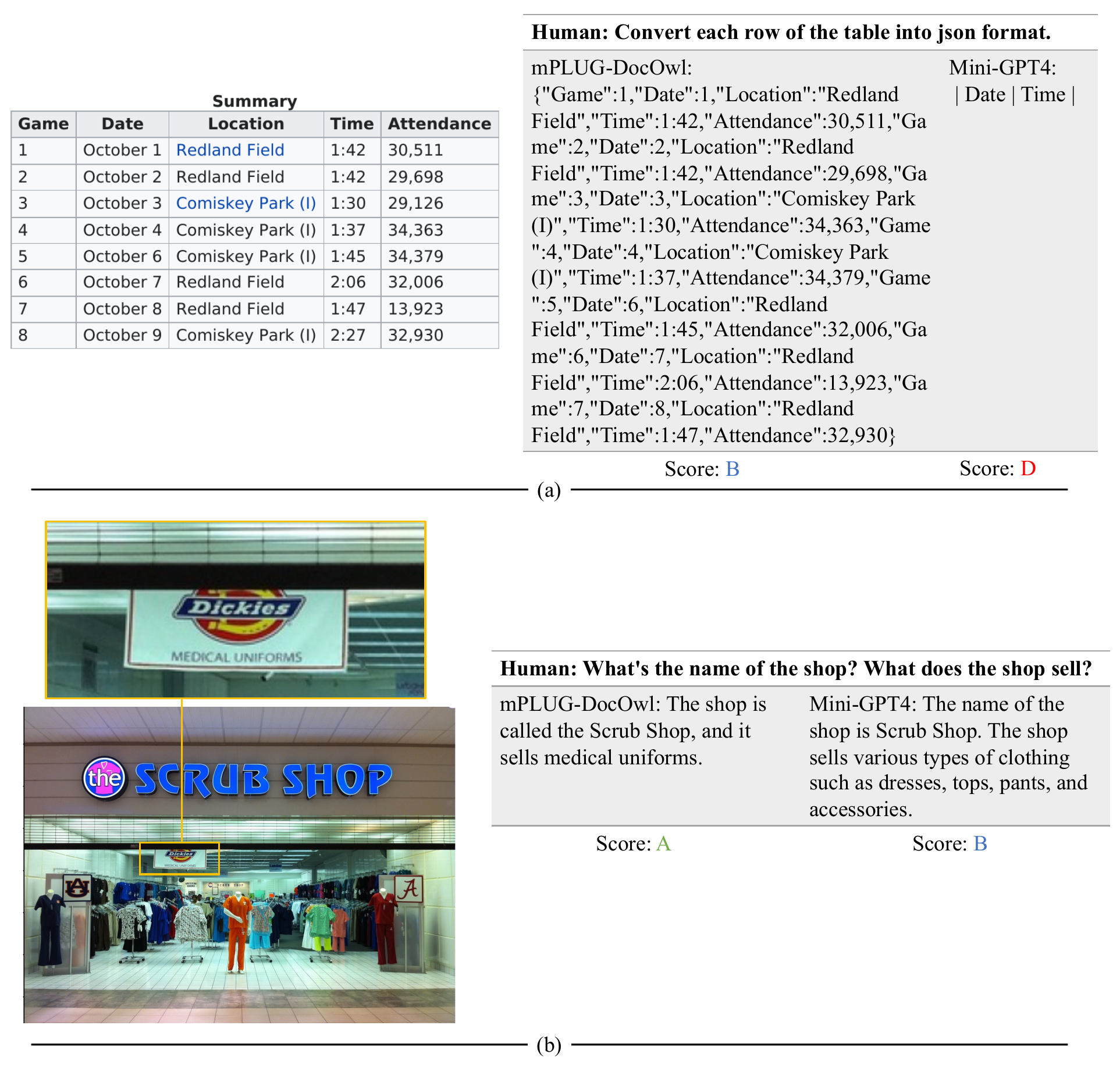}
    \caption{Qualitative comparison between \modelname and Mini-GPT4 on \evalname. Part one.}
    \label{fig:cases_human_1}
\end{figure}
\begin{figure}[tp]
    \centering
    \includegraphics[width=1.0\linewidth]{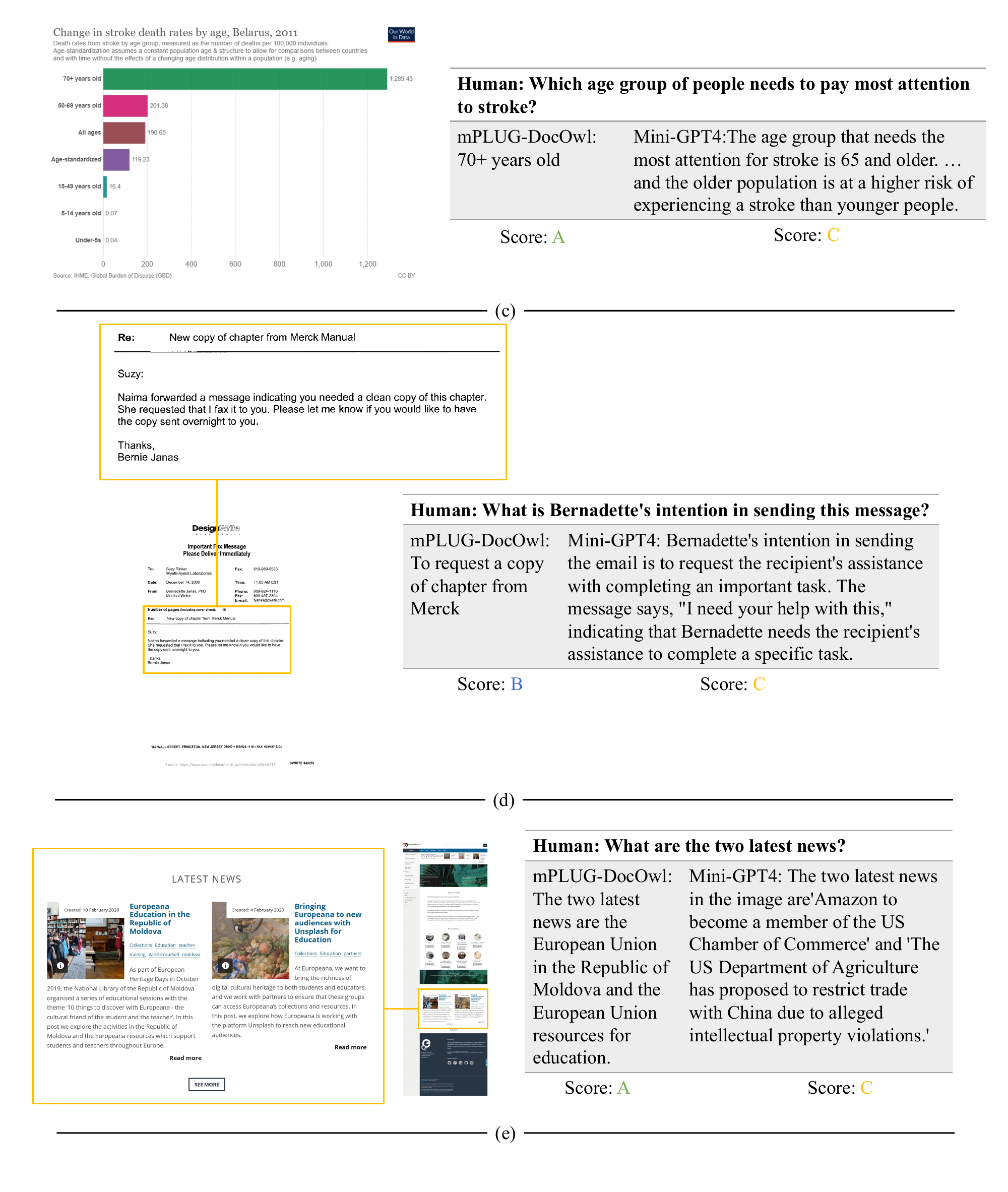}
    \caption{Qualitative comparison between \modelname and Mini-GPT4 on \evalname. Part two.}
    \label{fig:cases_human_2}
\end{figure}

\paragraph{\evalname Results} \Cref{fig:cases_human_1} and \Cref{fig:cases_human_2} present the comparison between \modelname and Mini-GPT4 on \evalname. \Cref{fig:cases_human_1} (a) requires models to convert a table into JSON format. Our \modelname correctly understands the instruction and return a string in JSON format, but misses the last row. Mini-GPT4 fails to comprehend the instruction and doesn't understand the content within the table. 
In \Cref{fig:cases_human_1} (b), both \modelname and Mini-GPT4 correctly recognize the name of the shop. 
However, Mini-GPT4 overlooks a smaller sign indicating clothes in this shop are medical uniforms. 
As for chart understanding in \Cref{fig:cases_human_2} (c), Mini-GPT4 gives a wrong answer and redundant response, while our \modelname gives a concise and correct response. In \Cref{fig:cases_human_2} (d), Bernadette's actual purpose is to confirm with Suzy if she would like to have the copy sent overnight. This not only requires the model to accurately recognize the text, but also to understand the relationships between involved persons. \modelname recognizes the phrase "request a copy of chapter," but misunderstands the subject and object. Mini-GPT4 only comprehends that this image is a mail scenario and provides a vague and hallucinatory response. In \Cref{fig:cases_human_2} (e), \modelname gives a correct summary of the two latest news but Mini-GPT4 generates  news irrelevant to the webpage screenshot.

\begin{figure}[tp]
    \centering
    \includegraphics[width=1.0\linewidth]{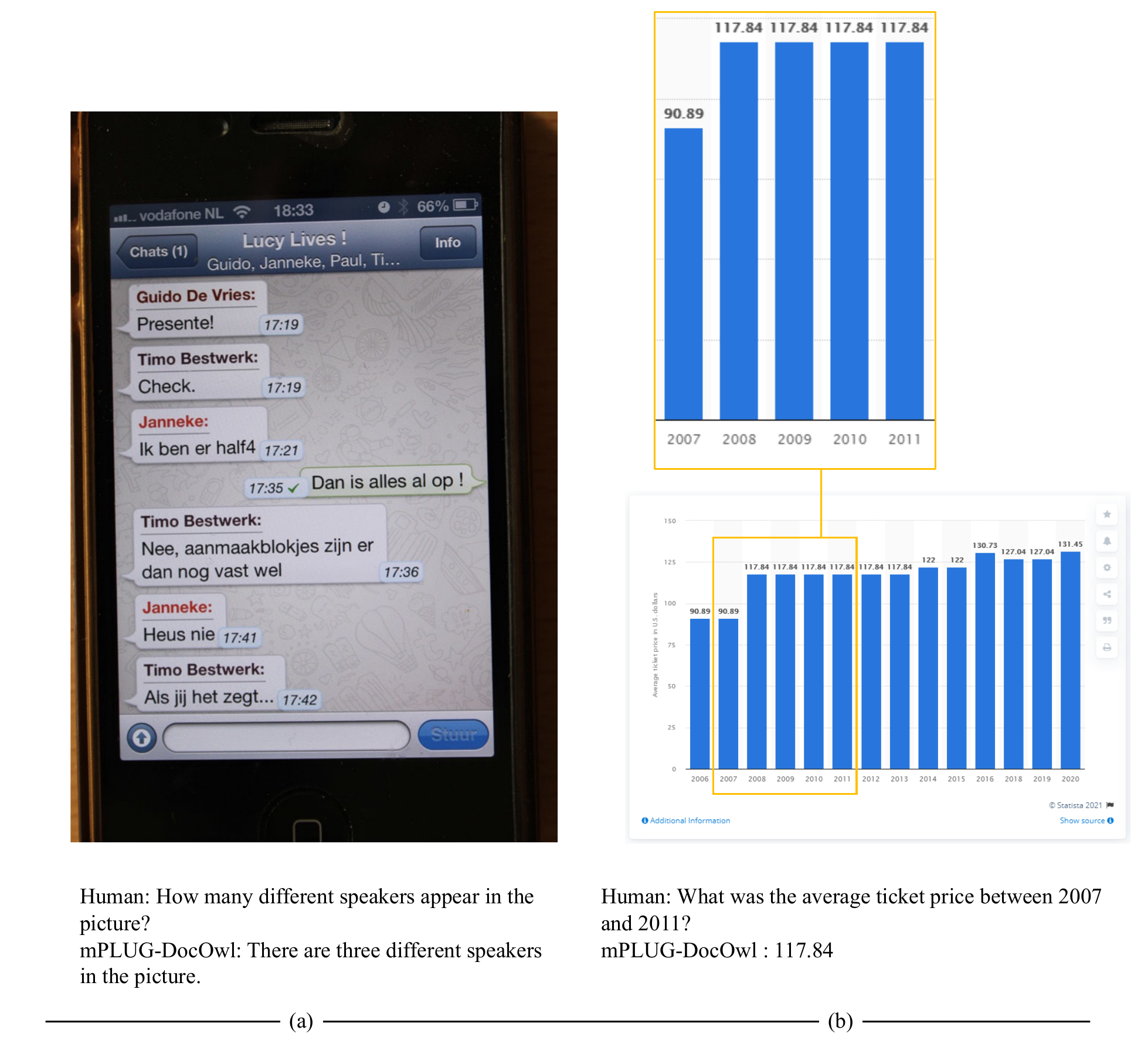}
    \caption{Failure cases on \evalname. Part one.}
    \label{fig:bad_case_1}
\end{figure}
\begin{figure}[tp]
    \centering
    \includegraphics[width=1.0\linewidth]{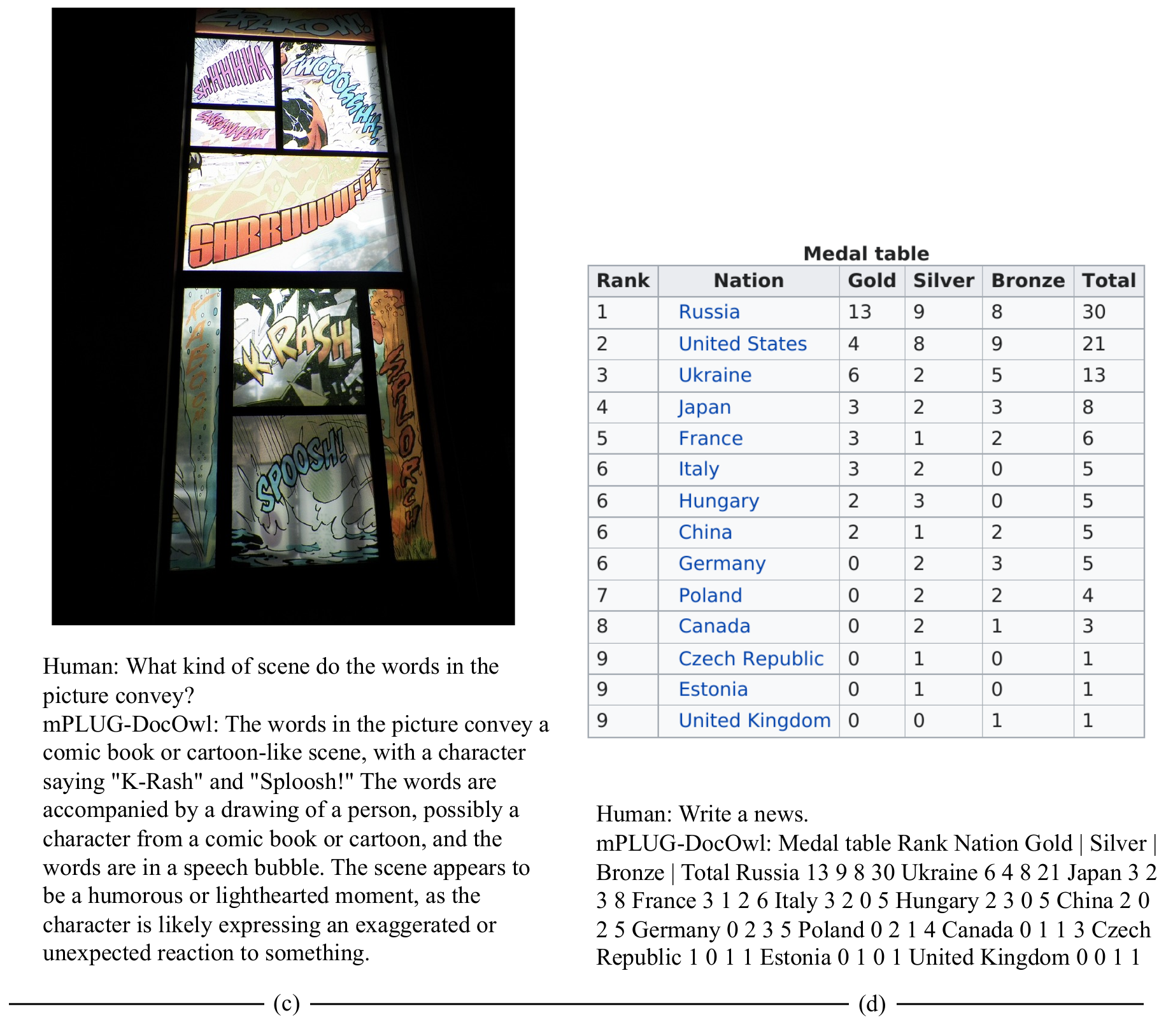}
    \caption{Failure cases on \evalname. Part two.}
    \label{fig:bad_case_2}
\end{figure}
The \evalname contains many challenging instruction understanding cases in the document domain. \Cref{fig:bad_case_1} and \Cref{fig:bad_case_2} show some wrong responses given by \modelname. In \Cref{fig:bad_case_1} (a), \modelname only takes note of the three names in the picture, but ignores the fact that the user itself is also a speaker. In \Cref{fig:bad_case_1} (b), \modelname fails to perform multi-step calculations on multiple elements in the image. In \Cref{fig:bad_case_2} (c), the model can understand the scene and the text in it, but fantasizes about non-existent characters. In \Cref{fig:bad_case_2} (d), \modelname fails to understand the instruction for writing news and only read the texts in  the tablet.

\section{Conclusion}
In this work, we infuse diverse ocr-free document understanding capabilities into mPLUG-Owl by incorporating document understanding data into instruction finetuning. Experiment results demonstrate that our \modelname achieves comparable or even better performance than existing OCR-free methods. Besides, benefiting from language-only and general vision-and-language instruction tuning, \modelname can better comprehend user instructions and intentions, enabling more complex interactions. Moreover, human evaluation on \evalname reveals that \modelname still struggles with  document-related commonsense reasoning, mathematical calculations, and creative generation. This provides valuable insights about developing stronger document understanding abilities with the LLM in the future.

\bibliographystyle{abbrvnat}
\clearpage
\bibliography{reference}

\clearpage

\end{document}